\documentclass[10pt,twocolumn,letterpaper]{article}

\usepackage[pagenumbers]{cvpr} 

\usepackage{graphicx}
\usepackage{amsmath}
\usepackage{amssymb}
\usepackage{booktabs}

\usepackage[pagebackref,breaklinks,colorlinks]{hyperref}

\usepackage[capitalize]{cleveref}
\crefname{section}{Sec.}{Secs.}
\Crefname{section}{Section}{Sections}
\Crefname{table}{Table}{Tables}
\crefname{table}{Tab.}{Tabs.}

\def\cvprPaperID{*****} 
\def\confName{CVPR}
\def\confYear{2022}

\begin{document}

\title{SDTracker: Synthetic Data Based Multi-Object Tracking}

\author{Yingda Guan$^{1}$, Zhengyang Feng$^{2}$ , Huiying Chang $^{3}$, Kuo Du$^{1}$, Tingting Li$^{4}$, Min Wang$^{1\dag}$ \\
\centerline{$^{1}$SenseTime Research \quad 
$^{2}$Shanghai Jiao Tong University\quad} \\
\centerline{$^{3}$Beijing University of Posts and Telecommunications \quad $^{4}$Beijing Forestry University \quad }\\
\centerline{\texttt{\{\small{guanyingda, dukuo, wangmin}\}\small{@sensetime.com}} }\\
\centerline{\texttt{\small{zyfeng97@sjtu.edu.cn,2018213064@bupt.edu.cn,litingting@bjfu.edu.cn}}}
}
\maketitle
\begin{abstract}
   We present SDTracker, a method that harnesses the potential of synthetic data for multi-object tracking of real-world scenes in a domain generalization and semi-supervised fashion. First, we use the ImageNet dataset as an auxiliary to randomize the style of synthetic data. With out-of-domain data,  we further enforce pyramid consistency loss across different "stylized" images from the same sample to learn domain invariant features. Second, we adopt the pseudo-labeling method to effectively utilize the unlabeled MOT17 training data. To obtain high-quality pseudo-labels, we apply proximal policy optimization (PPO2) algorithm to search confidence thresholds for each sequence. When using the unlabeled MOT17 training set, combined with the pure-motion tracking strategy upgraded via developed post-processing, we finally reach 61.4 HOTA.
\end{abstract}

\section{Introduction}
\label{sec:intro}


Pedestrian detection and tracking have a wide range of applications, such as self-driving and smart security.
Recent methods~\cite{zhang2021bytetrack} based on deep learning have achieved great performance. However, they require a large amount of labeled data, collecting and labeling such data is expensive and time-consuming, even violates privacy policies .
Therefore, synthesizing diverse large-scale datasets at a low cost is becoming a promising direction.
Recently, MOTSynth~\cite{fabbri2021motsynth} open sources the largest synthetic dataset for pedestrian detection and tracking, showing that models trained with synthetic data are  comparable to state-of-the-arts on tasks such as detection, ReID, and tracking.


The efficient use of synthetic data lies in bridging the synthetic-to-real gap. We address the problem by domain generalization, which aims to generalize a model trained on multiple source (\ie, synthetic) domains to distributionally different unseen (\ie, real) domains. In ~\cite{yue2019domain}, it is addressed by two key steps: domain randomization and consistency-enforced training. We extend this idea and use richer data augmentation to improve randomization. It is worth mentioning that MOTSynth~\cite{fabbri2021motsynth} confirms diversity matters. We argue that increasing the diversity of synthetic data is essential for enhancing domain randomization.
To further exploit the additional unlabeled MOT17 training set, we apply semi-supervised learning with the state-of-the-art pseudo-labeling approach. Inspired by \cite{wang2022pseudo}, we utilize proximal policy optimization to search for the best thresholds for pseudo labels.



Our MOT method is based on ByteTrack \cite{zhang2021bytetrack}. To better track real pedestrians with person detectors trained on synthetic data, we make several upgrades to ByteTrack's robustness, including motion compensations, full box estimation, confidence adjustments and a new post-processing pipeline to merge and interpolate tracks. Without any labeled real data, by the aid of our domain generalized detectors, our pure-motion MOT solution achieves $61.4$ HOTA on MOT17 test set, comparable to most of the state-of-the-arts trained on MOT17 training set.


\begin{figure*}[t]
	\centering
	\includegraphics[width=0.85\textwidth]{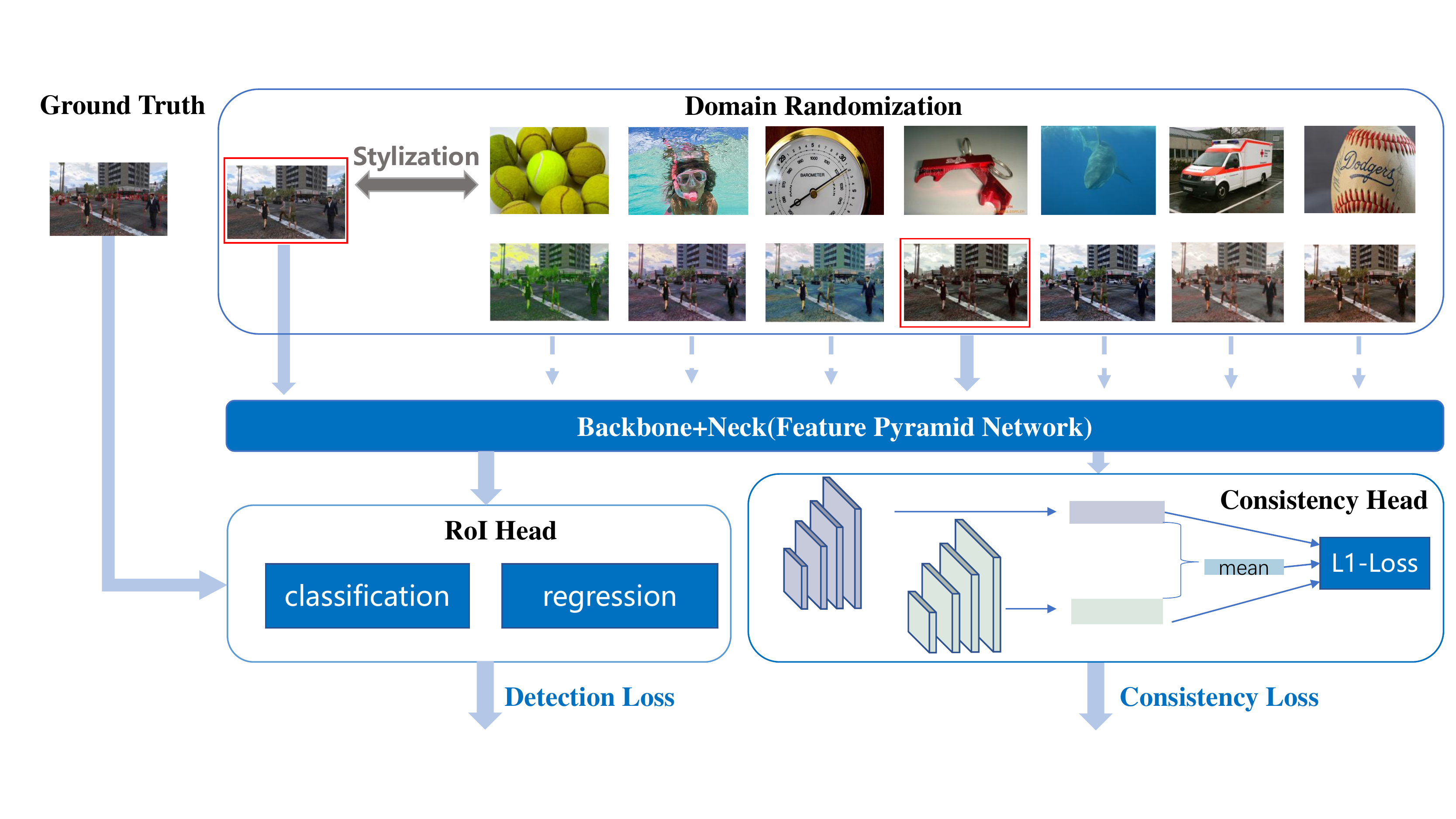}
	\vspace{-0.4in}
	\caption{The domain generalization framework consists of a domain randomization module and a consistency head.}
	\label{fig:pipeline}
\end{figure*}

\section{Method}
We employ the popular tracking-by-detection paradigm, which shows excellent performance on multi-object tracking. Our
pipeline consists of a person detector followed by a tracker module,
where the tracker associates person detection results through time into individual tracks.

\label{sec:method}

\subsection{Person Detector}
We apply domain generalization to the synthetic-to-real training regarding ~\cite{yue2019domain} when there is no access to real data. Semi-supervised learning based on ~\cite{wang2022pseudo} is utilized to further boost performance when real data was available. We re-implement the methods above in the detection framework with the following details.

\noindent \textbf{Domain Randomization with Stylization.}
 The first part of domain generalization is mapping the synthetic image to multiple auxiliary real domains when training. This ensures that the neural network can see sufficiently diverse data.  In current stage, as shown in ~\cref{fig:pipeline}, we select $K$  real-life categories from the ImageNet. Each category represents an auxiliary domain. We use CycleGan~\cite{zhu2017unpaired} to do style transfer for synthetic images, where cycle-loss can guarantee the generation of out-of-domain data while suppressing noise and distorted semantic information from source images. As a result, the training set is augmented in style by K times. During training, pairwise-sampling strategy is adopted. We select an original image and the corresponding stylized image  sampled randomly from $K$ domains for network forwarding. Classification and regression loss of the stylized image are also involved in backpropagation to improve the network's ability to detect on out-of-domain data. We chose $7$ categories in practice, which are great white shark, ambulance, barometer, tennis ball, can opener,  snorkel, and tennis ball.


\begin{equation}
\begin{split}
\mathcal{L}_{consis} &=\sum_{n, k} \sum_{l \in \mathcal{P}} \lambda_{l} L_{1}\left(\overline{g_{l}\left(\mathcal{P}_{n} \right)}, g_{l}\left(\mathcal{P}_{n}^{k}\right)\right)\\
&\overline{g_{l}\left(\mathcal{P}_{n} \right)}=\frac{1}{K+1} \sum_{k} g_{l}\left(\mathcal{P}_{n}^{l, k}\right)
\end{split}
 \label{eq:consis}
\end{equation}

\noindent \textbf{Consistency-Enforced Training.}
The second part of domain generalization is enforcing pyramid consistency across domains within an image to boost the generalization power of the model. Specifically, we employ a consistency head after the backbone, as shown in ~\cref{fig:pipeline}. Consider a set of images $\mathcal{I}_{n}=\left\{I_{n}^{k} \mid k=0,1, \ldots, K\right\}$  of $K + 1$ different styles with the same annotation, $\mathcal{P}$ denotes the feature pyramids, $\mathcal{P}_{n}^{l, k}$ is the feature map of input $\mathcal{I}_{n}^{k}$ at layer $l$. The consistency loss is shown in \cref{eq:consis}, where $g_{l}$ denotes AdaptiveAvgPooling. $L_{1}$ is $l_{1}$ distance. In our experiment, $\lambda_{l}$ is $0.5$. Directly 
employing the feature's distance of different domain images as consistency loss is strict, which is difficult to fitting. It is proven that using the mean as the target can help the network to learn the domain-invariant representation without affecting the optimization of the detection task. The consistency head is applied only during the training process and does not increase the inference time.

\noindent \textbf{Semi-Supervised Learning.}
To further improve the generalization of the model to the target domain, we use the model trained on synthetic data to generate pseudo-labels for the unlabeled MOT17 training set. The threshold value needs to be determined carefully to ensure the final performance. Motivated by ~\cite{wang2022pseudo}, we apply proximal policy optimization (PPO2) algorithm to search for thresholds that maximize the evaluation metric mAP. The training consists of $R$ rounds. In each round $r$, the PPO2 search process consists of $T$ sampling
steps. In each step, $M$ sets of parameters are sampled independently from a truncated normal distribution, the mean is initialized to $0.5$, and covariance  is fixed to $0.2$ in practice. $R$, $T$ and $M$ vary with the speed of model training. These sampled parameters are used
to train the detection model separately.
Then the mean of the distribution is updated by PPO2 algorithm according to the model's mAP. We search for each sequence separately due to different confidence distributions. To simplify the search process, the threshold is fixed for the whole schedule.

\noindent \textbf{Detection Ensemble.} To further enhance detection performance, we ensemble multiple models using Weighted Boxes Fusion (WBF) \cite{solovyev2021weighted}, which makes a weighted average of different detector's outputs instead of voting or selecting boxes. While we find that simply ensembling via NMS hurts performance, and difference between models are crucial for better ensemble results. We also apply test time augmentations (multi-scale testing and horizontal flip) for each of the models before ensemble.

\subsection{Tracker}

Same as ByteTrack \cite{zhang2021bytetrack}, our tracking method is purely motion-based. Preliminary experiments show that fusing appearance features like \cite{zhang2021fairmot} brings adverse effects.
However, a motion-based method highly relies on detection quality. In this contest, detectors are trained on synthetic data, we make several changes/upgrades to ByteTrack as follows.

\noindent \textbf{Full Box Estimation.}
We found that MOTSynth \cite{fabbri2021motsynth} detection annotation clipped out-of image border boxes. While the MOT17 benchmark use full human box.
Thus, we extend near border boxes by heuristics. For sequences $03$ and $04$, the videos are taken from a fixed bird eye's view parallel to the road, we fit linear functions to estimate a person's height $h$ as $h = ay_{t} + b$ and $h = ay_{b} + b$, where $y_t$ and $y_b$ are top and bottom Y coordinates of a bounding box. The data to fit these functions are extracted from one of our model's prediction on a random frame. For other sequences, simply extending boxes to a fixed aspect ratio ($2.6$) seems to suffice.

\noindent \textbf{Confidence Adjustments.}
ByteTrack \cite{zhang2021bytetrack} removes the confidence score from similarity scores for highly crowded scenes in MOT20. 
With synthetic data, we find our detectors tend to have lower or ill-calibrated scores. While this problem can somewhat be alleviated through hyper-parameter tuning, we find removing the confidence score for less crowded scenes in MOT17 helps to track.
We also raise the confidence score for near border detections, so they can participate in the first round of ByteTrack association, initiate a track, and confirm a newborn track.

\noindent \textbf{Motion Compensations.}
ByteTrack is cleanly based on motion, but motion estimating with the common Kalman filter is not robust enough. Inspired by StrongSORT \cite{du2022strongsort}, we apply the NSA Kalman filter and ECC camera motion compensation (estimated from MOT17 training set) for dynamic scenes, \ie, captured by a moving camera or featuring complex motion patterns. While for static scenes we do not use these techniques, since the vanilla Kalman filter runs well, these compensations would not help the performance.

\noindent \textbf{Better Interpolation.}
ByteTrack \cite{zhang2021bytetrack} applies interpolation to connect tracks with the same identity. However, ByteTrack still uses a very short track buffer and the interpolation is simply a linear function.
Concretely, ByteTrack usually terminates a track if it was lost for at most $30$ frames, which equals only $1$ second. To fully exploit the power of interpolation, we extend track buffer from $30$ up to $90$ for slow moving scenes. Furthermore, we operate the Gaussian-Smoothed Interpolation (GSI) from \cite{du2022strongsort} to generate a smooth trajectory, and remove very short tracks ($<10$ frames).

\noindent \textbf{Track Merge.}
Interpolation only connects within the same identity. However, some lost tracks encounter identity switches or re-appear as other identities. These problems can easily happen for an IoU-based tracker. Since the ReID features don't work well on synthetic data, and the only appearance-free linker AFLink \cite{du2022strongsort} requires training a link model on MOT17 annotations, yet only brings marginal improvement ($0.2$ HOTA in the paper). We design a merging process purely based on spatial and motion heuristics, which does not require training task-specific models like AFLink.
Specifically, for a lost track, we find the closest box by box center distance (after adjusting positions by estimated velocities). If this distance is lower than a fixed threshold, and the corresponding tracks satisfy certain similarity constraints, \eg, box area, lost time, these tracks are set to the same identity.
After that, another interpolation is applied to fill in the lost frames in a merged identity. This merging enables our submission to have a very high IDF1 score ($77.2$).
There is one downside for our merging policy: it only works with static scenes. How to design a similar process for dynamic scenes is an interesting future work direction.

\section{Experiments}
\noindent \textbf{Detection.}
To evaluate our detection model’s quality, we choose MOT17 training set as the validation set. Our codes are developed based on MMdetection~\cite{chen2019mmdetection}. We train four models with different capacities to further ensemble. Unlike the default settings of FPN~\cite{lin2017feature}, we add a small anchor set at scale $4$. Multi-scale training is adopted and the longer side is fix to $3000$. For the ResNet50~\cite{he2016deep}, SyncBN is used,  parameters trained in a self-supervised fashion~\cite{grill2020bootstrap} are employed as initialization weights, and the freezing of the first stage is removed. To handle domain gaps more robustly, all models are trained by default using multi types of data augmentation, including weather, blur, noise, and digital. For per image, the severity level and type are selected randomly. The final results are shown in  ~\cref{tab:detection}. Our experiments demonstrate the effectiveness of domain generalization and semi-supervised learning in synthetic-to-real gap.
Due to time and computing resource constraints, the detailed ablation experiments are not completed, which will be left for future work.

\noindent \textbf{Tracking.} We analyze our MOT designs in \cref{tab:track}. Here we use models before semi-supervised learning, which is proved to work well in \cref{tab:detection}. It can be seen that model ensemble, full box estimation, and interpolation increase performance on both static and dynamic scenes, motion compensation boost dynamic scene results by a large margin, while the proposed track merge technique can notably help static scenes. For all experiment configurations, we use grid search to find optimal hyper-parameters on MOT17 training set. The final configuration ($4$ models + FB + MC + GSI + TM) with semi-supervised learning corresponds to our $61.4$ HOTA submission on the test server.
\begin{table}[htb]
    \begin{center}
    \tabcolsep=2.0pt
    \begin{tabular}{@{}clcccccc}
    \toprule
    \# & \textbf{Algorithm} & \textbf{Backbone} & \textbf{Mstrain} & \textbf{DR} & \textbf{CT} & \textbf{SSL} & \textbf{mAP}\\
    \toprule
    0 & FPN & R50 & (800,1200) &  &  & & 44.8 \\
    \midrule
    1 & FPN & R50 & (800,1400) &\checkmark  &  & & 51.4 \\
    2 & FPN & R50 & (800,1200) & \checkmark &\checkmark & & 52.6 \\
    3 & Dyhead~\cite{dai2021dynamic} & swin-L~\cite{liu2021swin} & (800,1120) & \checkmark & & & 54.5 \\
    4 & FPN & swin-L & (800,1120) & \checkmark & & & 55.3 \\
    \midrule
    5 & FPN & R50 & (800,1400) & \checkmark &  &\checkmark & 54.8 \\
    6 & FPN & R50 & (800,1200) & \checkmark &\checkmark & \checkmark& 56.0 \\
    7 & Dyhead & swin-L & (800,1120) & \checkmark & &\checkmark & 55.7 \\
    8 & FPN & swin-L & (800,1120) & \checkmark & & \checkmark& 57.0 \\
    \bottomrule
    \end{tabular}
    \end{center}
    \vspace{-0.1in}
     \caption{Detection results evaluated on MOT17 training set. \textbf{Mstrain}: Multi-scale train. \textbf{DR}: Domain randomization with stylization. \textbf{CT}: Consistency-enforced training \textbf{SSL}: Semi-supervised learning.  \textbf{mAP}: mAP on MOT17 training set.}
    \label{tab:detection}
\end{table}
\begin{table}[ht]
    \centering
    \tabcolsep=2.0pt
    \begin{tabular}{lccccccc}
    \toprule
    \textbf{Models} & \textbf{FB} & \textbf{MC} & \textbf{GSI} & \textbf{TM} & \textbf{HOTA} & \textbf{HOTA-S} & \textbf{HOTA-D} \\
    \toprule
    (4) & &&&& 55.2 & 62.5 & 50.9 \\
    (1,2,3,4) & &&&& 55.9 & 64.7 & 52.6 \\
    (1,2,3,4) & && \checkmark && 56.6 & 65.0 & 53.9 \\
    (1,2,3,4) & & \checkmark & \checkmark && - & - & 56.6 \\  
    (1,2,3,4) & \checkmark & \checkmark & \checkmark && 60.5 & 69.5 & 57.3 \\
    (1,2,3,4) & \checkmark & \checkmark & \checkmark & \checkmark & - & 70.4 & - \\ 
    \bottomrule
    \end{tabular}
    \caption{MOT results evaluated on MOT17 training set. \textbf{FB}: full box estimation. \textbf{MC}: motion compensation. \textbf{GSI}: GSI interpolation. \textbf{TM}: Track merge. \textbf{HOTA}: HOTA on MOT17 training set. \textbf{HOTA-S}: HOTA on typical static sequence $04$. \textbf{HOTA-D}: HOTA on typical dynamic sequence $05$.
    }
    \label{tab:track}
\end{table}

\section{Conclusion}
This paper presents SDTracker, a novel approach to address synthetic-to-real training in multi-object tracking. Unlike previous domain adaption methods which require data from the target domain, the proposed approach designs a domain randomization module to generate diverse out-of-domain images from a single source domain. A consistency-based strategy is used to guide models to learn invariant representation across different domains. Proximal policy optimization additionally enables to obtain high-quality pseudo labels when using MOT17 training set.
SDTracker also makes several upgrades to the popular ByteTrack, increasing its robustness when using detectors trained with synthetic data.
Extensive experiments on MOT17 demonstrate that the method achieves state-of-the-art without labeled real data. 
\label{sec:formatting}




{\small
\bibliographystyle{ieee_fullname}
\bibliography{egbib}
}

\end{document}